\documentclass[letterpaper]{article} 
\usepackage{aaai25}  
\usepackage{times}  
\usepackage{helvet}  
\usepackage{courier}  
\usepackage[hyphens]{url}  
\usepackage{graphicx} 
\usepackage{natbib}  
\usepackage{caption} 
\usepackage{algorithm}
\usepackage{algorithmic}

\usepackage{booktabs}
\usepackage{multirow}
\usepackage{amsmath}
\usepackage{amssymb}
\usepackage{makecell}
\usepackage{subcaption}

\usepackage{newfloat}
\usepackage{listings}
\DeclareCaptionStyle{ruled}{labelfont=normalfont,labelsep=colon,strut=off} 
\floatstyle{ruled}
\newfloat{listing}{tb}{lst}{}
\floatname{listing}{Listing}
\title{CoPEFT: Fast Adaptation Framework for Multi-Agent Collaborative Perception with Parameter-Efficient Fine-Tuning}
\author{
    Quanmin Wei\textsuperscript{\rm 1, \rm 2}, Penglin Dai\textsuperscript{\rm 1, \rm 2}\thanks{Corresponding author. }, Wei Li\textsuperscript{\rm 1, \rm 2}, Bingyi Liu\textsuperscript{\rm 3}, Xiao Wu\textsuperscript{\rm 1, \rm 2}
}
\affiliations{
    \textsuperscript{\rm 1} School of Computing and Artificial Intelligence, Southwest Jiaotong University\\

    \textsuperscript{\rm 2} Engineering Research Center of Sustainable Urban Intelligent Transportation, Ministry of Education\\
    \textsuperscript{\rm 3} School of Computer Science and Artificial Intelligence, Wuhan University of Technology\\

    wqm@my.swjtu.edu.cn, penglindai@swjtu.edu.cn, liwei@swjtu.edu.cn, byliu@whut.edu.cn, wuxiaohk@gmail.com
}

\usepackage{bibentry}

\begin{document}

\maketitle

\begin{abstract}
Multi-agent collaborative perception is expected to significantly improve perception performance by overcoming the limitations of single-agent perception through exchanging complementary information. However, training a robust collaborative perception model requires collecting sufficient training data that covers all possible collaboration scenarios, which is impractical due to intolerable deployment costs. Hence, the trained model is not robust against new traffic scenarios with inconsistent data distribution and fundamentally restricts its real-world applicability. Further, existing methods, such as domain adaptation, have mitigated this issue by exposing the deployment data during the training stage but incur a high training cost, which is infeasible for resource-constrained agents. In this paper, we propose a Parameter-Efficient Fine-Tuning-based lightweight framework, CoPEFT, for fast adapting a trained collaborative perception model to new deployment environments under low-cost conditions. CoPEFT develops a Collaboration Adapter and Agent Prompt to perform macro-level and micro-level adaptations separately. Specifically, the Collaboration Adapter utilizes the inherent knowledge from training data and limited deployment data to adapt the feature map to new data distribution. The Agent Prompt further enhances the Collaboration Adapter by inserting fine-grained contextual information about the environment. Extensive experiments demonstrate that our CoPEFT surpasses existing methods with less than 1\% trainable parameters, proving the effectiveness and efficiency of our proposed method. 

\end{abstract}

%
\begin{links}
    \link{Code}{https://github.com/fengxueguiren/CoPEFT}
\end{links}

\section{Introduction}

Collaborative perception allows agents to share complementary information through communication, thereby enhancing a more comprehensive perception \cite{Han2023CollaborativePI}. This fundamentally becomes a new paradigm to overcome the long-standing limitations of single-agent perception, such as difficulties in distant and occluded perception \cite{Ren2022CollaborativePF}. Recent studies have highlighted the potential of collaborative perception in various realistic applications, including autonomous driving \cite{Wang2020V2VNetVC}, robot automation \cite{Li2022MultiRobotSC}, and UAV collaborative rescue \cite{hu2023aerial}. The field of collaborative perception is experiencing rapid growth, driven by the availability of high-quality datasets \cite{Xu2021OPV2VAO, xu2022v2x, Yu2022DAIRV2XAL}, the evolution of powerful fusion methods \cite{Chen2019CooperCP, Xu2021OPV2VAO}, and the development of robust collaborative systems \cite{wei2024asynchrony, li2024di}.

\begin{figure}[t]
\centering
\includegraphics[width=0.98\columnwidth]{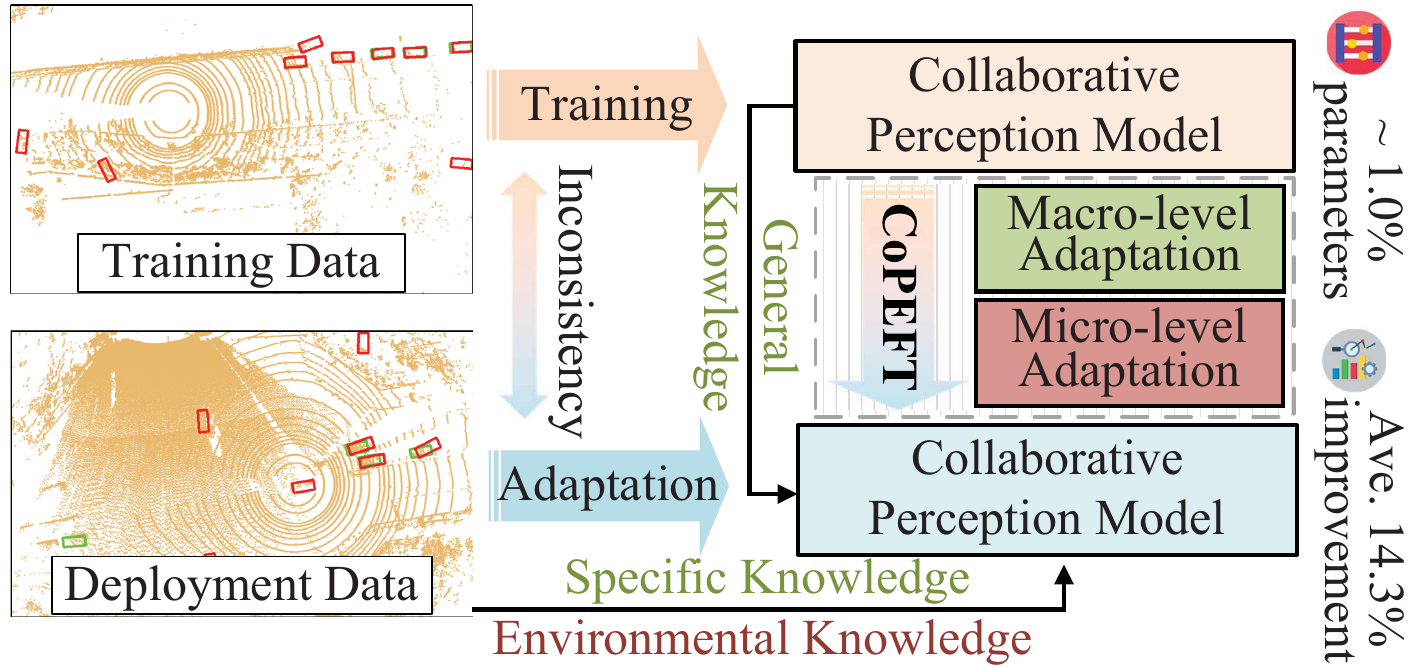} %
\caption{Illustration of CoPEFT. We mitigate the impact of inconsistent data distribution on collaborative perception by dynamically combining general knowledge derived from the training data with specific and environmental knowledge obtained from the deployment data. Here, general knowledge encompasses the general patterns of collaborative perception, specific knowledge represents the disparities between the new deployment and the training data, and environmental knowledge refers to fine-grained contextual information. Consequently, CoPEFT can fast adapt a well-trained model to various deployment environments at a low cost.}
\label{fig:fig1}
\end{figure}

Most existing collaborative perceptions \cite{Xu2021OPV2VAO, hu2024communication} rely on the assumption that the training and deployment data follow the same distribution, but ensuring this assumption is inherently challenging. In real-world deployment, the distribution inconsistency between the training and deployment data is a common occurrence \cite{yuan2023robust}, such as unseen road topology during training or different sensing patterns due to updated sensors, which often result in performance degradation of the trained collaborative perception model. The ineffective utilization of popular adaptation methods from other fields, such as transfer learning \cite{zhu2023transfer}, to address this dilemma is compounded by the collaborative nature and sparsity of the data in collaboration. Although some studies have explored collaborative domain adaptation by integrating deployment data during training \cite{li2023s2r, akong2023dusa}, they still require costly training from scratch for new data, which is unsuitable for resource-constrained agents. To address these issues, we aim to develop a unified and lightweight design that permits fast adaptation of the collaborative perception to inconsistent deployment environments while keeping the cost acceptable. Before that, we still need to answer the following questions. Firstly, given the inherent inconsistency in data distributions, \textit{how can we simultaneously preserve the shared patterns and unique characteristics in training and deployment data?} Secondly, it is still essential to identify the environmental context specific to agents for aiding the adaptation process. So \textit{how can we effectively guide collaborative perception in utilizing fine-grained environmental information?}

To address these questions, we propose a lightweight framework for collaborative perception, namely CoPEFT, which employs Parameter-Efficient Fine-Tuning for fast adaptation to new data distributions. The illustration of CoPEFT is presented in Figure \ref{fig:fig1}. The adaptation process within CoPEFT is structured into two levels to tackle the questions mentioned above separately.  From a macro perspective, a learnable Collaboration Adapter, with the assistance of sparse collaborative information, facilitates the dynamic combination of general knowledge from training data with specific knowledge related to deployment data, thereby adapting the feature map to the new distribution of deployment data. From a micro perspective, we develop an Agent Prompt that injects fine-grained environmental knowledge through a virtual agent, further enhancing the adaptation process. These two components jointly guide the trained collaborative perception model toward alignment with the deployment data, achieving significant performance gains while keeping costs at an acceptable level. In summary, CoPEFT seeks to equip collaborative perception with fast adaptation capabilities under limited supervision. 

CoPEFT has two significant advantages. Firstly, it is  resource-efficient, as it requires only a modest amount of labeled data and updates less than 1\% of the parameters. This feature enables the reuse of a trained model in different deployment environments without incurring expensive adaptations. Secondly, CoPEFT seamlessly integrates with existing collaborative perception systems, functioning as a plug-and-play universal plugin that effectively operates within intermediate and aggregated feature spaces. To validate the effectiveness of our CoPEFT, we conducted extensive experiments on three benchmark datasets for collaborative 3D object detection. The results consistently indicate that our method yields substantial improvements in performance. For instance, by adapting CoAlign \cite{Lu2022RobustC3} trained on OPV2V \cite{Xu2021OPV2VAO} to the DAIR-V2X \cite{Yu2022DAIRV2XAL} with a 10\% data availability rate using CoPEFT, we have doubled the performance at AP@70 compared to counterparts without adaptation or those trained from scratch. In comparison to the DUSA \cite{akong2023dusa} that updates all parameters for domain adaptation, CoPEFT improves the perception performance by 7.8\% at AP@70 while reducing the number of trainable parameters by 99\%. 

In summary, our contributions are three-fold. (1) To the best of our knowledge, this is the first comprehensive exploration of fast adaptation for collaborative perception, focusing specifically on alleviating the adverse effects of data inconsistency. (2) We propose a novel fast adaptation solution called CoPEFT, which can be seamlessly integrated with existing collaborative perception systems. It comprises two complementary components: a Collaboration Adapter for macro-level adaptation and an Agent Prompt for micro-level adaptation. (3) Extensive experiments on both simulated and real-world datasets demonstrate the superior performance of CoPEFT in comparison to SOTA methods.

\section{Related Work}

\subsection{Collaborative Perception}
Collaborative perception overcomes inherent limitations in single-agent perception by sharing complementary information among agents \cite{Han2023CollaborativePI}. Some early works can be categorized into early collaboration that shares raw observations \cite{Zhang2021EMPEM, Luo2023EdgeCooperNC} and late collaboration that transmits perception results \cite{ Miller2020CooperativePA}. However, these approaches often fail to strike a balance between communication efficiency and performance \cite{li2021learning}, hindering their practical applications. Recently, the intermediate collaboration paradigm, which operates in a compact feature space, has gained popularity as it offers a better performance-bandwidth trade-off \cite{Han2023CollaborativePI}.

V2VNet \cite{Wang2020V2VNetVC} represents a milestone in this field, employing a graph neural network to model the dynamic interactions among agents. After that, AttFuse \cite{Xu2021OPV2VAO} introduces the self-attention to aggregate intermediate features from different agents and release a high-quality OPV2V dataset. To alleviate the adverse impacts of pose errors, CoAlign \cite{Lu2022RobustC3} proposes an agent pose correction method and a multi-scale fusion method. To retain the advantages of early collaboration while reducing bandwidth, DiscoNet \cite{li2021learning} and MKD-Cooper \cite{li2023mkd} introduce knowledge distillation to guide the learning of the intermediate collaboration model. Most existing efforts assume that the training data for the collaborative perception model is comparable to the data encountered during deployment. However, this assumption is often deemed impractical in real-world deployment situations.


So far, only a few works, S2R-ViT \cite{li2023s2r} and DUSA \cite{akong2023dusa}, recognize the potential implications of this assumption not holding. These studies employ a technique called unsupervised domain adaptation, where labeled training data is combined with unlabeled deployment data during the training stage to uncover the distribution of the deployment data. They have the following limitations: (1) It is difficult to determine the deployment data during the training stage; (2) When the deployment data changes significantly, it requires costly re-training of the model from scratch; (3) This discriminative-based method may  distort the learned features \cite{tang2020unsupervised}, thereby affecting the final perceptual performance. In contrast to previous studies that focus solely on simulation-to-reality domain adaptation settings, our research not only addresses the aforementioned limitations but also broadens the applicability to a wider range of scenarios.

\subsection{Parameter-Efficient Fine-Tuning}
In natural language processing and computer vision, Parameter-Efficient Fine-Tuning (PEFT in short) offers an efficient alternative to full-parameter fine-tuning for specific tasks \cite{xin2024parameter}. The core idea behind PEFT is to achieve comparable performance to full-parameter fine-tuning by updating only a portion of the existing model's or newly added parameters. Inspired by the manually defined prompt \cite{petroni2019language}, the learnable prompt adjusts the model by adding a few parameterized input blocks into the input layer of the trained Transformer model \cite{jia2022visual, dong2022lpt, nie2023pro}. Some subsequent works have explored adjusting other elements of the Transformer architecture, such as attention block \cite{li2021prefix}. Another mainstream research is adapter, which inserts subnetworks containing bottlenecks within the backbone network to fine-tune the output of each layer \cite{houlsby2019parameter, chen2022adaptformer, xin2024vmt}. However, these works are all targeted at image or language models, which are not compatible with collaborative perception. In contrast, we introduce PEFT as a lightweight plugin that enables fast adaptation by encoding the inconsistency between the deployment and training data in collaborative perception.

In the context of collaborative perception, there is also a relevant method known as MACP \cite{ma2024macp}, which introduces the concept of PEFT to transfer a single-agent perception model to multi-agent perception. Different  from MACP, our goal is to achieve fast adaptation to new deployed scenario of collaborative perception with low cost by leveraging the complementary interaction of the proposed Collaboration Adapter and Agent Prompt

\begin{figure*}[t]
\centering
\includegraphics[width=1.9\columnwidth]{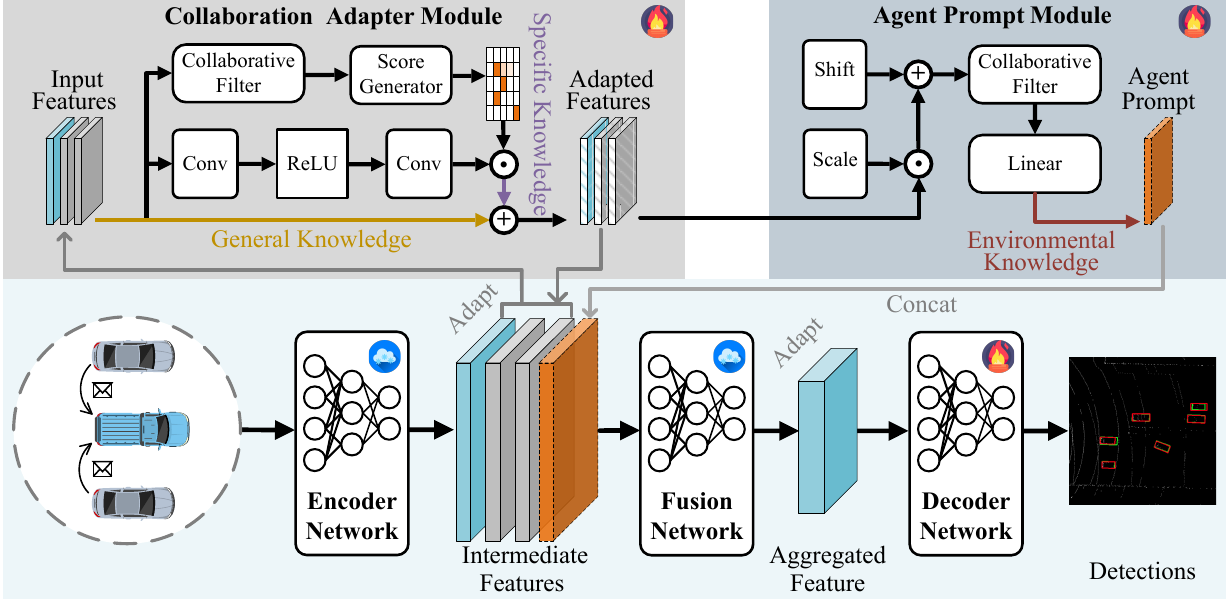} 
\caption{The overall architecture of CoPEFT. It involves standard components in intermediate collaboration augmented with two lightweight elements: a Collaboration Adapter and an Agent Prompt. (a) The Collaboration Adapter, guided by several collaborative perception priors, adapts the feature maps from a macro-level perspective for new data. (b) The Agent Prompt offers fine-grained environmental information from a micro-level perspective, which can be conceptualized as the insertion of a virtual agent to further assist in adapting feature maps. By updating only the parameters of the Collaboration Adapter, Agent Prompt, and Decoder Network, CoPEFT effectively realizes the dynamic combination of general, specific, and environmental knowledge for fast adaptation.}
\label{fig:overall_framework}
\end{figure*}

\section{Methodology}

\subsection{Overall Architecture}

Consider a set of $N$ agents, denoted as $\mathsf{A}=\{\mathsf{A}_1,\ldots,\mathsf{A}_N\}$, that are present in the current perceptual environment. Each agent is equipped with perceptual and computational capabilities. The goal is to encourage better 3D object detection through the cooperative sharing of complementary information among agents. Specifically, this paper focuses on intermediate collaboration that achieves a performance-bandwidth trade-off. For an ego agent $\mathsf{A}_i$ with local observation $\mathbf{O}_i$ and perception output $\mathbf{Y}_i$, the pipeline of our CoPEFT for collaborative 3D object detection is as follows

\begin{subequations} \label{eqa:overall}
    \begin{align}
    \mathbf{F}_i & =f_{\text {enc}}\left(\mathbf{O}_i\right) \label{eqa:overall_a}\\
    \widehat{\mathbf{F}}_i & =f_{\mathrm{c} \_ \text {ada1}}\left(\mathbf{F}_i\right), \mathbf{P}_i=f_{\mathrm{a} \_ \text {pro}}\left(\widehat{\mathbf{F}}_i\right), \label{eqa:overall_b}\\
    \mathbf{H}_i & =f_{\text {fus}}\left(\widehat{\mathbf{F}}_i,\left\{\widehat{\mathbf{F}}_j\right\}_{j \in \mathsf{A}, j\ne i }, \mathbf{P}_i\right), \label{eqa:overall_c}\\
    \widehat{\mathbf{H}}_i & =f_{\mathrm{c} \_ \text {ada2} }\left(\mathbf{H}_i\right)\label{eqa:overall_d}, \\
    \mathbf{Y}_i & =f_{\text {det}}\left(\widehat{\mathbf{H}}_i\right) \label{eqa:overall_e},
    \end{align}
\end{subequations}
where step \ref{eqa:overall_a} extracts intermediate BEV feature $\mathbf{F}_i$ from $\mathbf{O}_i$ using an Encoder Network $f_{\text {enc}}$, step \ref{eqa:overall_b} generates adapted feature $\widehat{\mathbf{F}}_i$ via a Collaboration Adapter $f_{\mathrm{c} \_ \text {ada1}}$ and fine-grained prompt $\mathbf{P}_i$ via an Agent Prompt module 
 $f_{\mathrm{a} \_ \text {pro}}$, step \ref{eqa:overall_c} merges intermediate features with a Fusion Network $f_{\text {fus}}$ to generates aggregated feature $\mathbf{H}_i$, step \ref{eqa:overall_d} adapts $\mathbf{H}_i$ using another Collaboration Adapter $f_{\mathrm{c} \_ \text {ada2}}$, and step \ref{eqa:overall_e} outputs the final detection results $\mathbf{Y}_i$ by a Decoder Network $f_{\text {det}}$. There are two aspects that require special attention. Firstly, in intermediate collaboration, each agent needs to standardize the coordinate system and send $\mathbf{F}_i$ after performing feature extraction (i.e., step \ref{eqa:overall_a}); and the remaining steps will be executed after receiving all data sent by other agents. Secondly, when steps \ref{eqa:overall_b} and \ref{eqa:overall_d} are removed, Equation \ref{eqa:overall} degenerates into the standard intermediate collaboration. This plug-and-play manner endows the CoPEFT with the flexibility to adapt to various collaborative perception systems.

Deploying a trained collaborative perception model directly in a new environment significantly increases the fatal risk due to potential inconsistency in data distribution. As shown in Figure \ref{fig:overall_framework}, after collecting a small amount of data with acceptable cost, we freeze most parameters and update only about 1\% of them (including the parameters of the Collaboration Adapter, the Agent Prompt, and the Decoder Network) to adapt to new environments. 

\subsection{Macro-level Adaptation via Collaboration Adapter}

Fast adapting collaborative perception model with new data poses a non-trivial problem under acceptable costs. On the one hand, updating only a small subset of parameters (e.g., Decoder Network) has a finite model's adaptability. On the other hand, insufficient data increases the risk of overfitting caused by noise. We note the homogeneity between the training and deployment data, which describe potentially general knowledge associated with collaborative perception, although significant differences may accompany them. Therefore, a potential solution to mitigate this dilemma is to dynamically combine general knowledge from training data with specific knowledge from limited deployment data. To concretely implement this idea, we propose the Collaboration Adapter from a macro perspective.

Specifically, the Collaboration Adapter $f_{\mathrm{c} \_ \text {ada}}:\mathbb{R} ^ {N \times D} \rightarrow \mathbb{R} ^ {N \times D}$ is a lightweight network for fast adaptation under limited supervision signals, where $D$ denotes the feature dimension. As depicted in subgraph at the upper left corner of Figure \ref{fig:overall_framework}, BEV feature $\mathbf{F}_i \in \mathbb{R} ^ {N \times D}$ is first transformed via a convolutional adapter. Distinct from conventional adapter methods \cite{houlsby2019parameter, chen2022adaptformer}, we adopt convolutional layers $\{\operatorname{Conv}_{\text{up}}, \operatorname{Conv}_{\text {down}}\}$ with a default bottleneck rate of 4 instead of linear ones to match sparse BEV inputs. With $\mathbf{F}_i$ as the input, the Collaboration Adapter $f_{\mathrm{c} \_ \text {ada}}$ can be expressed as follows

\begin{equation}
    \begin{aligned} 
    \widehat{\mathbf{F}}_i & =f_{\mathrm{c} \_ \text {ada}}\left(\mathbf{F}_i\right), \\ 
    & =\underbrace{\mathbf{F}_i}_{\text{general knowledge}} \oplus \underbrace{S \odot \operatorname{Conv}_{\text{up}} \sigma\left(\operatorname{Conv}_{\text {down}} \mathbf{F}_i\right)}_{\text{specific knowledge}},
    \end{aligned}
    \label{eq:eq2}
\end{equation}
where $\oplus$ denotes the element-wise addition, $\odot$ denotes the element-wise multiplication, and $\sigma$ is the ReLU \cite{nair2010rectified}. $S$ is a modulation score containing the priors from collaboration, which will be described in detail below.

Furthermore, the significance of inter-agent interaction and foreground confidence information is undeniable in collaborative perception. To leverage these valuable priors, we add a parallel branch into standard adapter architecture, consisting of a Collaborative Filter $ \operatorname{ColF}:\mathbb{R} ^ {N \times D} \rightarrow \mathbb{R} ^ {1 \times D}$ and a Score \ Generator $\operatorname{ScoG}:\mathbb{R} ^ {1 \times D} \rightarrow \mathbb{R} ^ {1 \times D}$ in series to obtain a modulation score $S \in \mathbb{R} ^ {1 \times D}$. Specifically, these two components are implemented using parameter-free max pooling $\operatorname{Max}$ and convolutional layer $\operatorname{Conv}_{1\times1}$ with kernal size $1\times1$, formulated as   

\begin{equation}
    \begin{aligned}
    S & = \operatorname{ScoG}(\operatorname{ColF}(\mathbf{F}_i)), \\
      & = \operatorname{Conv}_{1 \times 1}(\operatorname{Max}(\mathbf{F}_i)).
    \end{aligned}
    \label{eq:eq3}
\end{equation}

In the standard CoPEFT, the aggregated feature $\mathbf{H}_i$ is adapted using another non-sharing Collaboration Adapter. The macro-level adaptation can refer to Equations \ref{eq:eq2} and \ref{eq:eq3}.

\subsection{Micro-level Adaptation via Agent Prompt}

The Collaboration Adapter inherently provides global adaptation, referred to as macro-level adaptation, that is shared among arbitrary inputs. However, it falls short in capturing the fine-grained information in collaborative perception, where different agents occupy distinct environments. To overcome this limitation, we propose the concept of Agent Prompt, which enhances the adaptation capability from a micro-level perspective. Agent Prompt is derived from a learnable prompt but offers several notable features. Unlike the existing prompts \cite{houlsby2019parameter, xin2024vmt}, which are typically randomly initialized and concatenated with the embeddings of other input blocks to collectively serve as the input of the Transformer layer, this improved Agent Prompt aligns with our design goal. Specifically, it is initialized with the output of the Collaboration Adapter, enabling awareness of the input instance. Furthermore, it extends into the general intermediate feature space to accommodate diverse collaborative perception systems.

\begin{table*}[t]

    \setlength{\tabcolsep}{1mm}
    
    \centering
    \begin{tabular}{c|c|c|c|c}
    
    \toprule
        Method & Publication & Parameter & AP@50 & AP@70 \\ 
    \midrule
        None & - & 0/12,896,384 & 0.429 & 0.217 \\ 
    
        Training from Scratch & - & 12,896,384/12,896,384 & 0.423 & 0.210 \\ 
        Decoder Network only & - & 5,140/12,901,524 & 0.515 & 0.276 \\ 
        SSF \cite{lian2022scaling} & NeurIPS 2022 & 5,780/12,902,164 & 0.518 & 0.280 \\ 
        Adapter \cite{houlsby2019parameter, chen2022adaptformer} & ICML 2019, NeurIPS 2022 & 42,420/12,938,804 & 0.579 & 0.380 \\ 

        MACP \cite{ma2024macp} & WACV 2024 & 43,060/12,939,444 & \underline{0.597} & \underline{0.389} \\ 
        DUSA \cite{akong2023dusa} & ACM MM 2023 & 14,213,266/14,213,266 & 0.514 & 0.340 \\ 
    \midrule
        CoPEFT (Ours) & AAAI 2025 & 111,270/13,007,654 & \textbf{0.610} & \textbf{0.418} \\ 
    \bottomrule

    \end{tabular}
        \caption{Collaborative 3D object detection on the DAIR-V2X dataset, where the shared base model, CoAlign \cite{Lu2022RobustC3}, is trained on the OPV2V dataset. None denotes the direct deployment of the collaborative perception model to new scenarios without any adaptation. The optimal and sub-optimal performances are highlighted in \textbf{bold} and \underline{underline}, respectively.}
    \label{tab:tab1}
\end{table*}

\begin{table*}[t]

    \centering
    \begin{tabular}{c|c|c|c|c|c|c|c|c}
    \toprule
    \multicolumn{1}{c}{} & \multicolumn{2}{c}{\textbf{1\%}} & \multicolumn{2}{c}{\textbf{2\%}} & \multicolumn{2}{c}{\textbf{5\%}} & \multicolumn{2}{c}{\textbf{20\%}} \\
    \midrule
        Method & AP@50 & AP@70 & AP@50 & AP@70 & AP@50 & AP@70 & AP@50 & AP@70 \\ 
    \midrule
        None & 0.429 & 0.217 & 0.429 & 0.217 & 0.429 & 0.217 & 0.429 & 0.217 \\ 
        Training from Scratch & 0.139 & 0.053 & 0.199 & 0.069 & 0.332 & 0.137 & 0.599 & 0.395 \\ 
        Decoder Network only & 0.432 & 0.183 & 0.469 & 0.228 & 0.497 & 0.240 & 0.532 & 0.292 \\ 
        SSF & \textbf{0.507} & \underline{0.221} & \underline{0.513} & 0.264 & 0.532 & 0.285 & 0.567 & 0.333 \\ 
        Adapter & 0.315 & 0.089 & 0.466 & 0.199 & \underline{0.575} & 0.352 & 0.619 & 0.409 \\ 

        MACP & \underline{0.455} & 0.192 & \underline{0.513} & 0.239 & \underline{0.575} & \underline{0.362} & \underline{0.623} & \underline{0.414} \\ 
    \midrule
        CoPEFT (Ours) & \textbf{0.507} & \textbf{0.268} & \textbf{0.517} & \textbf{0.302} & \textbf{0.596} & \textbf{0.384} & \textbf{0.627} & \textbf{0.434} \\ 
    \bottomrule
    \end{tabular}
        \caption{Collaborative 3D object detection under different data availability rates.}
    \label{tab:tab2}
\end{table*}

As shown in the subgraph at the upper right corner of Figure 2, the Agent Prompt module $f_{\mathrm{a} \_ \text {pro}}:\mathbb{R} ^ {N \times D} \rightarrow \mathbb{R} ^ {1 \times D}$ initially employs the parameter-efficient SST \cite{perez2018film, liu2023generalized} to modulate the output $\widehat{\mathbf{F}}_i$ of the Collaboration Adapter $f_{\mathrm{c} \_ \text {ada1}}$, thereby generating environmental context information $\mathbf{E}_i \in \mathbb{R} ^ {N \times D}$: 

\begin{equation}
    \mathbf{E}_i  = \operatorname{Scale}\odot  \widehat{\mathbf{F}}_i \oplus \operatorname{Shift},
\end{equation}
where $\operatorname{Scale} \in \mathbb{R} ^ {C}$ and $\operatorname{Shift} \in \mathbb{R} ^ {C}$ are scaling and shifting operator for transformation, and $C$ is channel. Next, it passes a $\operatorname{ColF}$ and enters a linear layer $\operatorname{Linear}$ to improve the expressive capabilities. This process yields an Agent Prompt $\mathbf{P}_i \in \mathbb{R} ^ {1 \times D}$ that matches size of a single intermediate feature $\mathbf{F}_i$ to provide fine-grained environmental knowledge:

\begin{equation}
    \mathbf{P}_i  = \underbrace{\operatorname{Linear}(\operatorname{ColF}(\mathbf{E}_i))}_{\text{environmental knowledge}}. 
\end{equation}

Finally, the Agent Prompt $\mathbf{P}_i$ is concatenated with the existing adapted features $\{\widehat{\mathbf{F}}_i,\left\{\widehat{\mathbf{F}}_j\right\}_{j \in \mathsf{A}, j\ne i }\}$ as the input $\mathbf{I}_i \in \mathbb{R} ^ {N+1 \times D}$ to the Fusion Network $f_{\text {fus}}$. Intuitively, the insertion of the Agent Prompt can be seen as a virtual agent $\mathsf{A}_{N+1}$ participating in the collaborative perception process. Its fusion interaction with the Fusion Network enhances the complementary adaptation for the Collaboration Adapter.

\section{Experiments}

\subsection{Dataset}

We conduct extensive experiments on three public benchmark datasets for multi-agent collaborative perception: OPV2V \cite{Xu2021OPV2VAO}, DAIR-V2X \cite{Yu2022DAIRV2XAL}, and V2XSet \cite{xu2022v2x} datasets. Specifically, the OPV2V dataset, a large-scale simulation dataset designed to simulate vehicle-to-vehicle interactions, comprises 11K frames of data and 232K 3D bounding boxes. Each frame includes an average of 3 agents, ranging from a minimum of 2 to a maximum of 7. The V2XSet is a vehicle-to-everything simulation dataset. Similar to the OPV2V dataset, it is jointly collected from the high-fidelity simulators CARLA \cite{dosovitskiy2017carla} and OpenCDA \cite{xu2021opencda}. It contains 11K frames of data from both the intelligent vehicle and intelligent infrastructure perspectives. Finally, the DAIR-V2X dataset is the first vehicle-to-everything real dataset. DAIR-V2X is more challenging compared to other simulation datasets due to inevitable noise. We follow existing works \cite{Lu2022RobustC3, li2024di} and employ supplementary 3D annotations for DAIR-V2X.

\subsection{Experimental Setup}

\subsubsection{Evaluation Metrics.}
We select collaborative 3D object detection accuracy as the experimental evaluation metric. We fix the evaluation area as $x \in [-100m, 100m]$ and $y \in [-40m, 40m]$ for all datasets, thereby excluding objects outside this spatial range. The experimental results are quantified using Average Precisions (AP) at Intersection-over-Union (IoU) thresholds of 50 and 70, denoted as AP@50 and AP@70, respectively.

\subsubsection{Settings and Implementation Details.}
To evaluate the effectiveness of our CoPEFT in fast adaptation for collaborative perception under low-cost conditions, we consider collaborative 3D object detection using a small amount of labeled data available with a default proportion set at 10\%. This setting is reasonable as the annotation process of small data can be completed with acceptable manual effort and time, aided by automatic annotation tools. Since our method is universal for any collaborative perception method, by default, we employ a multi-scale fusion-based CoAlign method \cite{Lu2022RobustC3} as the base model to concretely implement CoPEFT. Additionally, to further evaluate the flexibility of CoPEFT, we also use two other collaborative perception models, AttFuse \cite{Xu2021OPV2VAO} and MKD-Cooper \cite{li2023mkd}, and vary the availability rates of the data, specifically 1\%, 2\%, 5\%, and 20\%. In contrast, the unsupervised domain adaptation method (DUSA) requires 100\% unlabeled deployment data to achieve competitive results.

We first train the collaborative perception models using their default settings to simulate the trained models. Then, we update the parameters of the Collaboration Adapter, the Agent Prompt, and the Decoder Network by utilizing partially available deployment data while keeping the parameters of the backbone network frozen. CoPEFT is optimized by the Adam optimizer with a learning rate of 0.002 and a batch size of 2. The maximum epoch for all methods is fixed at 20. Since the proportion of trainable parameters of CoPEFT is less than 1\%, the adaptation time can saved by tens of times compared to some traditional domain adaptation methods (e.g., DUSA). All experiments are implemented with PyTorch on an NVIDIA 3090 GPU.

\begin{table}[t]

    \centering
    \begin{tabular}{c|c|c|c}
    \toprule
       Fusion & Method & AP@50 & AP@70 \\ 
    \midrule
        \multirow{6}{*}{AttFuse} &  None & 0.442 & 0.203 \\ 
        &Training from Scratch & 0.326 & 0.167 \\ 
        &Adapter & \underline{0.552} & \underline{0.359} \\ 
        &MACP & 0.533 & 0.350 \\ 
        &DUSA & 0.457 & 0.324 \\ 
        &CoPEFT (Ours) & \textbf{0.554} & \textbf{0.374} \\ 
    \midrule
        \multirow{6}{*}{\makecell[c]{MKD-\\Cooper}} & None & 0.320 & 0.157 \\ 
        &Training from Scratch & 0.480 & 0.300 \\ 
        &Adapter & \underline{0.548} & \underline{0.348} \\ 
        &MACP & 0.537 & 0.339 \\
        &DUSA & 0.460 & 0.319 \\ 
        &CoPEFT (Ours) & \textbf{0.554} & \textbf{0.362} \\ 
    \bottomrule
    \end{tabular}
        \caption{Collaborative 3D object detection using AttFuse \cite{Xu2021OPV2VAO} and MKD-Cooper \cite{li2023mkd} as base model.}
    \label{tab:tab3}
\end{table}

\subsection{Quantitative Evaluation} 

As shown in Table \ref{tab:tab1} and Table \ref{tab:tab2}, our CoPEFT, which has less than 1\% trainable parameters, outperforms all baseline methods, including PEFT methods developed for other fields, PEFT, and domain adaptation methods tailored for collaborative perception. For instance, when adapting with only 10\% of the deployment data, CoPEFT improves the detection performance by an average of 19.1\% compared to the unadapted baseline, is 8.7\% higher than the domain adaptation method DUSA for collaborative perception, and demonstrates promising improvements compared to PEFT methods for other fields. Several crucial observations can be revealed from these tables. Firstly, due to the substantial inconsistency between the training and deployment data, the collaborative perception model suffers from severe performance degradation, as demonstrated by the performance in the "None" rows. Secondly, training a model from scratch for a new environment requires a large amount of labeled data and excessive training costs. Thirdly, existing PEFT methods fail to achieve satisfactory adaptation performance relative to ours, potentially because they lack alignment with collaborative perception. Finally, although the DUSA for domain adaptation does not require labeled data, its performance boost is not substantial and relies on a large amount of deployment data. Consequently, our CoPEFT offers a novel strategy for collaborative perception to adapt to the new deployment environment under low-cost conditions, particularly in terms of low training and data costs, which correspond to a small number of trainable parameters and a minimal amount of labeled data, respectively. 

\begin{table}[t]

    \centering
    \begin{tabular}{c|c|c}
    \toprule
        Method & AP@50 & AP@70  \\ 
    \midrule
        None & 0.918 & 0.839  \\ 
        Training from Scratch & 0.871 & 0.699  \\ 
        Adapter & 0.928 & 0.849  \\ 
         
        MACP & \underline{0.930} & \underline{0.851}  \\
        DUSA & 0.889 & 0.842  \\
    \midrule
        CoPEFT (Ours) & \textbf{0.933} & \textbf{0.854} \\ 
    \bottomrule
    \end{tabular}
        \caption{Collaborative 3D object detection is adapted from the OPV2V to the V2XSet dataset.}
    \label{tab:tab4}
\end{table}

\begin{table}[t]
    \centering

    \begin{tabular}{c|c|c|c}
    \toprule
        \makecell[c]{Collaboration\\ Adapter} & \makecell[c]{Agent\\ Prompt} & AP@50 & AP@70 \\ 
    \midrule
        - & - & 0.429 & 0.217 \\ 
        \checkmark & - & 0.604 & 0.408 \\ 
        - & \checkmark & 0.578 & 0.372 \\ 
        \checkmark & \checkmark & \textbf{0.610} & \textbf{0.418} \\ 
    \bottomrule
    \end{tabular}
        \caption{Ablation study on main components.}
    \label{tab:tab5}
\end{table}

To validate the flexibility of CoPEFT, we report the results using alternative collaborative perception models as the base model in Table \ref{tab:tab3}. Note that the table only includes the competitive comparison methods, and the subsequent results are presented in a similar format. It is apparent that these models are not robust against changes in the deployment environment, while CoPEFT endows them with the capability to adapt. Furthermore, the experimental results under relatively similar training and deployment distributions are detailed in Table \ref{tab:tab4}, where the two datasets adopt a simulation collection scheme with comparable configurations. Despite minor performance degradation in collaborative perception models due to slight distributional differences, CoPEFT consistently delivers performance enhancements.

\subsection{Qualitative Evaluation} 

\begin{figure}[t]
\centering
\includegraphics[width=0.98\columnwidth]{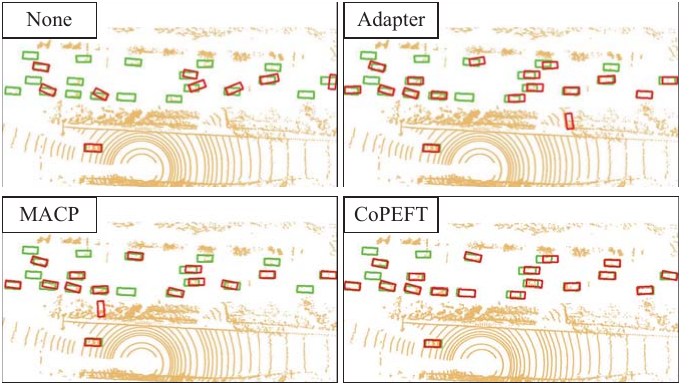} 
\caption{Qualitative comparison. The green and red 3D bounding boxes represent ground truth and prediction, respectively. Best viewed in color.}
\label{fig:qualitative_comparison}
\end{figure}

To intuitively illustrate the superiority of CoPEFT, we present the results of qualitative comparison in Figure \ref{fig:qualitative_comparison}. We can observe that CoPEFT achieves better results in 3D object detection. Specifically, CoPEFT has made significant contributions in several aspects, including reducing false positives, enhancing true positives, and improving matching accuracy. These results are consistent with the quantitative evaluation mentioned above, proving that CoPEFT can effectively eliminate the impact of inconsistency between training and deployment data on collaborative perception.

\begin{table}[t]

    \centering
    \begin{tabular}{c|c|c|c|c}
    \toprule
        \makecell[c]{Convo- \\lution}& \makecell[c]{Collabora- \\tive Filter} & \makecell[c]{Score \\ Generator} & AP@50 & AP@70 \\ 
    \midrule
        - & - & - & 0.579 & 0.380 \\ 
        \checkmark & - & - & 0.588 & 0.389 \\ 
        - & \checkmark & - & 0.588 & 0.392 \\ 
        - & - & \checkmark & 0.594 & 0.395 \\ 
        - & \checkmark & \checkmark & 0.589 & 0.397 \\ 
        \checkmark & \checkmark & - & 0.583 & 0.386 \\ 
        \checkmark & \checkmark & \checkmark & \textbf{0.604} & \textbf{0.408} \\ 
    \bottomrule
    \end{tabular}
        \caption{Ablation study on Collaboration Adapter.}
    \label{tab:tab6}
\end{table}

\subsection{Ablation Study}

\subsubsection{Effectiveness of Main Components.} We check the effectiveness of each component and report the ablation results in Table \ref{tab:tab5}. The first row represents the performance without adaptation, while the second and third rows indicate the performance using only a specific component. The final row represents the complete CoPEFT. Introducing the Collaboration Adapter or Agent Prompt achieves substantial performance improvement compared to the base model without adaptation. Notably, Collaboration Adapters contribute an additional 3.6\% improvement at AP@70 over the individual Agent Prompt. We further combine the two components, resulting in an average performance boost of 19.1\% over the base model. These results highlight the effectiveness of each design within CoPEFT.

\subsubsection{Internal Components of Collaboration Adapter and Agent Prompt.} As shown in Tables \ref{tab:tab6} and \ref{tab:tab7}, additional ablation experiments are conducted to gradually incorporate internal designs into the Collaboration Adapter and Agent Prompt. These tables follow a similar organization, where the first row corresponds to the naive PEFT method developed for other domains, and the last row represents the complete module. Both complete modules within CoPEFT surpass the baseline and their respective ablation counterparts in performance. Specifically, the Collaboration Adapter and Agent Prompt achieve average improvements of 2.6\% and 5.8\% over the baseline, respectively. Furthermore, they demonstrate varying degrees of enhancement for the incomplete ablation variants. Thus, the experimental results unquestionably validate the effectiveness of integrating collaborative perception priors into the two elements of CoPEFT.

\subsubsection{Analysis on CoPEFT variants.} 
\begin{figure}[t]
\centering
\includegraphics[width=0.82\columnwidth]{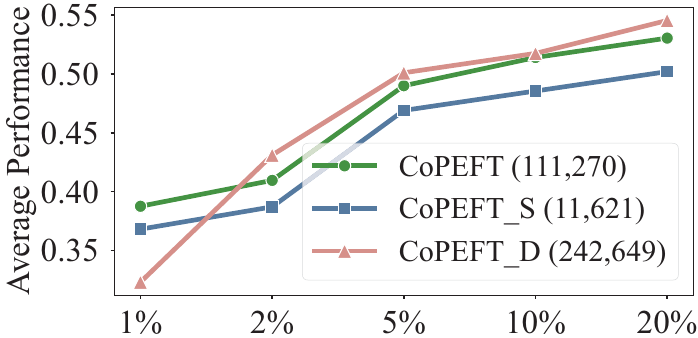} %
\caption{Comparisons of CoPEFT variations.}
\label{fig:fig_CoPEFT_variant}
\end{figure}

We analyze the effect of incorporating our CoPEFT into various positions. The specifics of the variants, including parameters and performances, are outlined in Figure \ref{fig:fig_CoPEFT_variant}. Besides the standard CoPEFT that applies to the intermediate and aggregation feature space, we also introduce a more lightweight version, CoPEFT\_S, which only adapts intermediate features, and a powerful CoPEFT\_D, which inserts extra Collaboration Adapters into each layer of the Fusion Network. Note that none of these three variants have additional bandwidth requirements as they operate on the ego agent. The results demonstrate that CoPEFT\_S significantly reduces the number of tunable parameters, yet does not offer any performance advantage. In addition, CoPEFT\_D yields satisfactory results when more deployment data is available, but with roughly double parameters compared to CoPEFT. Therefore, considering the trade-off between training cost and performance, we prefer the standard CoPEFT.

\begin{table}[t]
  
    \centering
    \begin{tabular}{c|c|c|c}
    \toprule
        Instance-aware & \makecell[c]{Collaborative \\Filter} & AP@50 & AP@70 \\ 
    \midrule
        - & - & 0.526 & 0.308 \\ 
        \checkmark & - & 0.572 & 0.360 \\ 
        \checkmark & \checkmark & \textbf{0.578} & \textbf{0.372} \\ 
    \bottomrule
    \end{tabular}
      \caption{Ablation study on Agent Prompt.}
    \label{tab:tab7}
\end{table}

\section{Conclusion}

In this paper, we investigate the performance degradation of collaborative perception in inconsistent deployment data with training data. We propose a general framework, called CoPEFT, to fast adapt collaborative perception models to the new deployment environment under acceptable costs. This framework consists of the Collaboration Adapter and Agent Prompt. The Collaboration Adapter focuses on macro-level adaptation by aligning feature maps with the distribution of the deployment data. Conversely, the Agent Prompt pursues micro-level adaptation by incorporating fine-grained environmental information. CoPEFT only updates less than 1\% of parameters with the cooperation of the two components, rendering it an efficient and effective solution.
\section*{Acknowledgements}
This work was partially supported by the National Natural Science Foundation of China under Grant Numbers 62172342, 62372387 and 62001400, the Natural Science Foundation of Hebei Province, China under Grant Number 2022105003, Key R\&D Program of Guangxi Zhuang Autonomous Region, China (Grant No. AB22080038, AB22080039), Sichuan Science and Technology Program (Grant No. 2024NSFSC0494), China Postdoctoral Science Foundation (Grant No. 2020T130547, No.2021M702713) and Fundamental Research Funds for the Central Universities (2682024ZTPY044).

\bibliography{aaai2025}

\end{document}